# Multi-chain Graph Refinement and Selection for Reliable Reasoning in Large Language Models

Yujiao Yang    Jing Lian    Linhui Li
Dalian University of Technology
Dalian, China
yjyang@mail.dlut.edu.cn, lianjing@dlut.edu.cn, lilinhui@dlut.edu.cn

**Abstract**

*The complex reasoning ability of Large Language Models (LLMs) poses a critical bottleneck for their practical applications. Test-time expansion methods such as Tree-of-Thought (ToT) and Graph-of-Thought (GoT) enhance reasoning by introducing intermediate reasoning structures, tree search, or graph-based exploration mechanisms. However, their reasoning strategies suffer from limited diversity, redundant search branches, and inadequate integration and error correction across heterogeneous reasoning paths. To address these limitations, we propose a novel reasoning framework called Multi-chain Graph Refinement & Selection (MGRS), which first generates multiple diverse reasoning trajectories for a given problem, refines candidate responses using a composite self- and cross-verification strategy, then constructs a reasoning relation graph and estimates the success rate of intermediate nodes, and finally computes cumulative success rates to select the most reliable answer and corresponding reasoning trajectory. Experimental results demonstrate that MGRS significantly advances both the reasoning capability and computational efficiency of reasoning enhancement methods. Across six benchmark datasets spanning four distinct tasks, MGRS achieves an average accuracy of 82.9%, outperforming state-of-the-art baselines by a clear margin of 2.1%. Remarkably, on the 24-point game, MGRS attains 100% accuracy for the first time, while delivering a 13.6× speed-up compared to the leading Forest of Thoughts framework.*

## 1. Introduction

Since the establishment of the large-scale pre-training paradigm, Large Language Models (LLMs) have have exhibited unprecedented generalization capabilities not only in traditional NLP tasks such as machine translation, and conversational generation, but also in emerging fields such as code generation, AI4Science, and embodied intelligence. However, when confronted with complex problems that require logical reasoning, the conventional next-token prediction mechanism exhibits inherent limitations, including systematic biases and error accumulation [1]. To alleviate these problems, the Chain-of-Thought (CoT) framework [2] introduces reasoning instructions or exemplars with intermediate steps into prompts, thereby guiding models to perform multi-step reasoning during inference and generate final answers through a structured deductive process. This approach has yielded substantial performance improvements on mathematical, commonsense, and symbolic reasoning benchmarks.

Although CoT provides a viable paradigm for complex reasoning, its single-chain, unidirectional generation mechanism remains limited in terms of reasoning depth, breadth, and accuracy. Several studies have therefore proposed various extensions to mitigate these shortcomings. Among them, Tree-of-Thought (ToT) [3] and Graph-of-Thought (GoT) [4] replace the linear chain with a multi-branch tree or a directed acyclic graph, enabling global optimization through concurrent exploration and backtracking. Self-Consistency [5] and Majority Voting [6] sample multiple independent reasoning chains to estimate the most consistent outcome. Moreover, Reflexion [7] and Self-Correct [8] introduce a "generate–criticize–revise" closed loop that iteratively refines reasoning trajectories and corrects intermediate errors.

The aforementioned extensions provide possible solutions for enhancing model performance. However, several challenges remain unresolved. First, although traditional tree- and graph-based reasoning frameworks are theoretically capable of modeling branching structures and complex dependencies, they often lack mechanisms that guide LLMs to expand branches at critical decision points. As a result, these frameworks tend to degenerate into ensembles of multiple, highly similar reasoning chains. Second, not all nodes necessarily require branching. From the perspective of human reasoning, it is more natural to explore multiple solution paths for a given problem—each corresponding to a distinct combination of branches within the reasoning tree. The diversity of these combinations is often sufficient for problem-solving and cross-verification, while also helping to reduce computational overhead. Furthermore, branch expansion in existing approaches typically relies on end-level voting across independent chains, overlooking step-wise error propagation. Even high-confidence segments may accumulate biases during intermediate steps. Therefore, it is essential to improve the mechanism of chain-of-thought enhancement and effectively adapt it to tree- or graph-based reasoning frameworks, thereby enhancing overall reasoning accuracy.

This paper proposes a novel reasoning framework—Multi-chain Graph Refinement & Selection (MGRS)—that enhances the reasoning capabilities of large language models (LLMs) by reforming the inference process. MGRS generates multiple diverse reasoning chains for a given problem and constructs a directed acyclic graph (DAG) through dependency analysis and similarity-based node merging. To further strengthen reasoning, we design a composite verification and refinement strategy that integrates self-verification within individual reasoning chains, cross-verification among branches, and targeted error correction. Moreover, we introduce an answer selection strategy based on estimated success probability, which identifies the most reliable final answer—rather than the most frequent reasoning chain—by jointly considering success likelihood and voting consistency. Experimental results demonstrate that the proposed MGRS framework substantially improves the reasoning performance of LLMs, enabling them to address complex tasks with higher accuracy and efficiency.

## 2. Related Work

### 2.1. Structured reasoning optimization

Structured reasoning methods explicitly organize the reasoning process into structured forms such as chains, trees, or graphs, enabling large language models (LLMs) to generate intermediate steps in a stepwise manner according to predefined logic.

In chain-based approaches, Chain-of-Thought (CoT) [2] decomposes a problem into a sequence of relatively simple intermediate steps and gradually reasons toward the final conclusion. By reducing the overall difficulty of reasoning, this approach effectively improves accuracy on complex mathematical and logical tasks. Subsequent research has enhanced CoT in terms of reliability and zero-shot triggering efficiency. For instance, TiM [9] stores intermediate reasoning steps in external memory slots and introduces backtracking for long-chain reasoning. Zero-Shot-CoT [10] achieves accuracy comparable to traditional CoT prompting using only a prefix template, eliminating the need for in-context examples. Plan-and-Solve Prompting [11] first outlines a reasoning plan and then executes it step by step, mitigating issues of missing steps and computational errors in zero-shot CoT reasoning.

Although CoT performs well in several tasks, it still exhibits limitations when applied to scenarios requiring multidimensional and nonlinear reasoning. To enhance reasoning flexibility, a series of more structured and search-capable paradigms have been proposed. Among them, Tree of Thoughts (ToT) [3, 12] generates multiple candidate reasoning paths at each step and employs breadth-first or depth-first search combined with self-evaluation mechanisms to backtrack to any parent node. Subsequent studies such as MCTSr [13] integrate Monte Carlo Tree Search (MCTS) [14], while Everything-of-Thoughts [15] and Forest-of-Thought (FoT) [16] incorporate input augmentation and multi-sampling strategies to further enhance performance. Nevertheless, existing tree-based methods remain limited in achieving efficient and diverse search. In contrast, our approach leverages a group of differentiated reasoning chains to effectively cover the major branches of the reasoning tree, and introduces reverse reasoning chains within these branches, thereby achieving diverse exploration while substantially improving computational efficiency.

On the other hand, Graph of Thoughts (GoT) [4] organizes reasoning nodes into a directed acyclic graph (DAG), allowing for merging, looping, and skipping between arbitrary nodes to achieve a fully connected reasoning topology. Building upon this idea, several works have demonstrated its effectiveness[17, 18]. Atom of Thoughts (AoT) [19] further decomposes the reasoning process into atomic cognitive entities consisting of a subproblem and its corresponding local reasoning outcome. The sequential and mutually exclusive relationships among these units are represented through dependency edges, thereby forming an extensible graph-based reasoning structure. However, in this framework, the DAG construction relies on a single atomic decomposition process (i.e., a single reasoning chain), which inherently lacks diversity and completeness, and consequently provides no effective mechanism for validation or correction. In contrast, our approach enhances reasoning diversity through differentiated reasoning chains, improves stability via multi-sampling of key reasoning paths, and integrates comprehensive verification and answer selection, thereby substantially improving overall reasoning performance.

### 2.2. Controlled reasoning optimization

Controlled reasoning methods incorporate search, sampling, or verification mechanisms to dynamically regulate the reasoning trajectories and output behaviors of large language models (LLMs) during inference. In this domain, multi-sampling and self-consistency are commonly adopted [5, 20, 21], where the model generates multiple reasoning paths and aggregates their outputs via majority voting or consistency-based scoring to mitigate the noise of single-sample generation. Similarly, several studies explore path re-ranking and answer aggregation strategies to identify more reliable outputs from diverse reasoning candidates [22, 23].

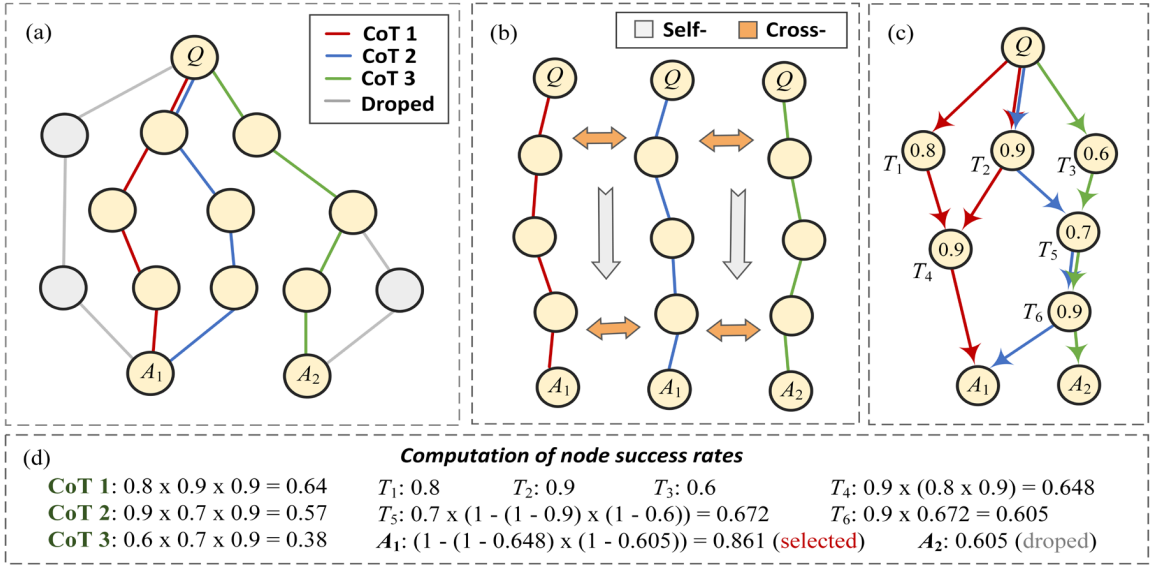

Figure. 1. The workflow of MGRS framework. Given an input question $Q$, MGRS first generates multiple diverse reasoning trajectories that yield preliminary answers A1 and A2, then refines these trajectories and answers through a composite self- and cross-verification strategy. Next, it constructs a Directed Acyclic Graph (DAG) based on the refined reasoning paths and estimates the single-step success rate of each intermediate node. Finally, it computes the cumulative success rate of each answer and reasoning trajectory, and selects the answer and reasoning trajectory with the highest cumulative success rate.

On the other hand, verification and editing approaches aim to mitigate error propagation by introducing verification procedures at each step or several key stages of the reasoning process. Representative efforts include Verify-and-Edit [24], which leverages external knowledge for post-hoc verification and correction, and Deductive Verification [25], which performs stepwise logical checks during generation to improve reasoning reliability. The verification mechanism can be implemented as an internal self-checking strategy within the model, or as an external module that integrates knowledge bases, program executors, or dedicated discriminators, thereby reducing factual and computational errors to a certain extent and improve the overall stability of multi-step inference procedure.

Although the aforementioned methods effectively improve reasoning accuracy, they are primarily designed for chain-based reasoning architectures. In contrast, our approach adapts these mechanisms to the characteristics of multi-branch reasoning, thereby extending their applicability to tree- and graph-based frameworks. Specifically, for multi-sampling and self-consistency, we repeatedly sample key differentiated branches and select reliable answers through an answer selection mechanism based on estimated success rates and voting confidence. For verification and refinement, we develop a composite strategy tailored to branched structures, integrating internal verification, cross-verification, and content refinement. These designs substantially enhance the accuracy and robustness of the overall reasoning process.

## 3. Methodology

This section introduces the algorithmic implementation of MGRS. The overall workflow of MGRS is illustrated in Figure 1, which comprises four key stages: (1) differentiated reasoning chains generation, (2) composite verification and refinement, (3) reasoning relation graph construction, and (4) answer and strategy selection.

### 3.1. Differentiated reasoning chains generation

To initiate the MGRS reasoning process, we first use LLMs to perform multiple differentiated reasoning attempts for a given problem, in order to generate a diverse set of step-by-step reasoning strategies $\Gamma$:

$$\Gamma = \left\{T^{(1)}, T^{(2)}, ..., T^{(M)}\right\} \quad (1)$$

Where $T^{(i)}(1 < i < M)$ denotes a differentiated reasoning chain, whose inference process can be formalized as a probabilistic sampling procedure:

$$A^{(i)} \sim p\left(A \mid T^{(i)}, Q, P^{(i)}\right) \prod_{j=0}^{N} p\left(T_j^{(i)} \mid T_{<j}^{(i)}, Q, P^{(i)}\right) \quad (2)$$

Here, $A^{(i)}$ represents the reasoning outcome of the i-th chain, and $T_j^{(i)}$ denotes its j-th reasoning step, which depends on all previous steps $T_{<j}^{(i)}$, the question $Q$, and the corresponding prompt $P^{(i)}$. The $P^{(i)}$ consists of a standard multi-step reasoning prompt and a differentiated prompt component. The latter can take the form of an abstracted summary of prior methods, an automatically generated heuristic, or a predefined theoretical or methodological overview, guiding the model to explore the problem from distinct reasoning perspectives.

To improve stability, the first $m$ branches $(1 < m < M)$ are repeatedly sampled $N$ times. Each sampling is performed independently under the same prompt. For each branch, we extract the logits corresponding to its final answer field and compute a perplexity-based confidence score for the k-th sampled result $S^{(k)}$:

$$S^{(k)} = \exp\left(-\frac{1}{L}\sum_{l=0}^{L}\log p(x_l | x_{<l})\right) \quad (3)$$

A lower perplexity score indicates higher model confidence in the generated answer. For each set of K samples from the i-th branch, we select several reasoning chains with low $S^{(k)}$ value as the representative $T^{(i)}$. It is worth noting that differentiated reasoning chains encompass both forward and reverse reasoning trajectories. The application of reverse reasoning chains is further discussed in Section 4.2.

### 3.2. Composite verification and refinement

Self-correction, as a reasoning enhancement mechanism aligned with human cognitive processes, has demonstrated strong effectiveness in prior works such as Self-Refine [26]. In MGRS, we extend this mechanism to multi-branch reasoning structures and propose a composite verification and refinement framework that integrates intra-chain self-verification and inter-chain cross-verification. In the intra-chain verification stage, the LLM iteratively examines each reasoning chain for logical consistency and computational accuracy, correcting any detected mathematical or inferential errors. The cross-verification stage then evaluates the consistency of final outcomes across branches; when discrepancies are observed, the framework identifies the points of divergence and refines the erroneous segments accordingly. Both verification and refinement processes are primarily guided by LLM-driven prompting, while incorporating a set of predefined validation and correction rules (e.g., arithmetic validation logic for basic operations). Through this composite mechanism, MGRS enables the model to effectively detect and repair reasoning errors, thereby enhancing the overall accuracy and robustness of the reasoning process.

### 3.3. Reasoning relation graph construction

In this stage, we construct a reasoning relation graph based on the verified differentiated reasoning chains to structurally organize and consolidate the reasoning process. Specifically, the LLM analyzes the dependency relationships among sub-steps across all reasoning chains and constructs a directed acyclic graph (DAG) that explicitly represents the logical flow of reasoning. This structured representation unifies reasoning paths with overlapping content and captures multi-path dependencies as well as shared reasoning evidence, thereby enabling more interpretable and reliable answer selection in subsequent stages.

Within the workflow, similar sub-steps from different reasoning chains are first merged to obtain $n$ distinct sub-steps together with their associated dependency relationships. Each sub-step's success probability is then independently estimated. The directed acyclic graph can thus be defined as follows:

$$G = (S, E), S = \{S_i\}_{i=1}^{n}, E \subseteq S \times S \quad (4)$$

Where each node $S_i$ represents a merged sub-step, and its weight $W_i$ denotes the estimated success probability of that sub-step, $0 < W_i < 1$. Each directed edge $(S_i, S_j)$ indicates that $S_j$ provides the necessary information for solving $S_i$.

Through this LLM-based analysis, the original linear dependency structure of chain-based reasoning is transformed into a hierarchical and interconnected dependency graph, in which the merged sub-steps naturally inherit the dependencies of their original counterparts.

### 3.4. Answer and reasoning strategy selection

The constructed directed acyclic graph (DAG) may yield multiple candidate answers. In MGRS, we select the answer with the highest success rate as the final output and identify its corresponding reasoning path—the one with the highest cumulative success rate—as the explanation chain. For a single reasoning chain with n steps $\{S_1, S_2, ..., S_n\}$ and corresponding single step success rate $\{W_1, W_2, ..., W_n\}$, the cumulative success rate is computed as the product of the success rates at each step:

$$P_{CoT} = \prod_{i=1}^{n} S_i \quad (5)$$

For the reasoning graph $G$, the reasoning branches for a given question $Q$ may converge to identical or distinct answers. We therefore compute the cumulative success rate of each node and propagate it layer by layer to obtain the cumulative success rate of all answer nodes. For a source node without predecessors, its cumulative success rate equals its own success rate: $P(S_i) = W_i$. For a node with a single parent $S_j$, its cumulative success rate depends on that of its parent: $P(S_i) = W_i \times P(S_j)$. For a node with n parent nodes $\{S_{j1}, S_{j2}, ..., S_{jn}\}$, the reasoning succeeds if any of its parents succeed; thus, this process is modeled using the Noisy-OR aggregation mechanism:

$$P(S_i) = W_i \times \left[1 - \prod_{k=1}^{n}\left(1 - P(S_{jk})\right)\right] \quad (6)$$

Table 1: Performance Comparison Across Tasks (%). Results are reported as exact match accuracy for MATH, GSM8K, BBH, and MMLU-CF, and F1 scores for HotpotQA and LongBench.

| Method | MATH | GSM8K | BBH | MMLU-CF | HotpotQA | LongBench | Avg. |
|---|---|---|---|---|---|---|---|
| CoT | 78.3 | 90.9 | 78.3 | 69.6 | 67.2 | 57.6 | 73.7 |
| CoT-SC(n=5) | 81.8 | 92.0 | 83.4 | 71.1 | 66.2 | 58.6 | 75.5 |
| Self-refine | 78.7 | 91.7 | 80.0 | 69.7 | 68.3 | 58.2 | 74.4 |
| AFlow | 83.0 | 93.5 | 76.0 | 69.5 | 73.5 | 61.0 | 76.1 |
| FoT(n=8) | 82.5 | 94.0 | 82.4 | 70.6 | 66.7 | 59.1 | 75.9 |
| AoT | 83.6 | 95.0 | 86.0 | 70.9 | 80.6 | 68.5 | 80.8 |
| **MGRS (Ours)** | **86.8** | **96.5** | **86.2** | **74.2** | **83.5** | **69.9** | **82.9** |

Note that as illustrated in Figure 1(c), there are cases where a node has multiple parent nodes originating from the same reasoning chain. In such situations, the nodes with the same origin are merged into a single node, whose success rate is computed using the product formula (Equation 5) and then used in the calculation of Equation 6 as a merged node.

By iteratively applying the above rules in the topological order of the graph, we obtain the cumulative success rate for each answer node. The final answer and its corresponding reasoning path are selected as those with the highest cumulative success rate.

## 4. Experiments

We conduct a comprehensive series of experiments across multiple widely used benchmarks to thoroughly evaluate the performance of MGRS. Specifically, we first perform systematic testing on six benchmarks spanning four types of reasoning tasks using different backbone models, and compare the results with state-of-the-art methods. Subsequently, we evaluate MGRS on the 24-point game dataset—a representative benchmark for tree-search-based LLM reasoning frameworks—against comparable approaches. Finally, we perform ablation studies to assess the contribution of key design components.

Experimental results demonstrate that MGRS substantially enhances both the reasoning capability and computational efficiency of large language models, achieving an average accuracy of 82.9% across six benchmark datasets covering three task categories, outperforming state-of-the-art baselines by 2.1%. Moreover, MGRS achieves 100% accuracy on the 24-point game for the first time, accompanied by a 13.6× speed-up over the Forest of Thoughts framework. The ablation studies further confirm the critical roles of Differentiated Reasoning Chains, Composite Verification and Refinement, Reasoning Relation Graph, and Success-Rate-Based Optimal Path Search in contributing to the overall performance improvement.

### 4.1. Main experiments

**Benchmark Description.** In the main benchmark evaluation, we implemented the MGRS on the GPT-4o-mini reasoning model to balance reasoning capability and computational efficiency. To comprehensively assess the performance of MGRS, we selected four representative categories of reasoning tasks. For mathematical reasoning, we employed the MATH dataset [27] and GSM8K [28] to evaluate the model's ability in symbolic reasoning and numerical computation. For logical reasoning, we used the multiple-choice subset of the BBH dataset [29] to assess the model's performance in formal logical judgment. For knowledge-intensive reasoning, we adopted the MMLU-CF dataset [30] to evaluate the model's understanding and application of knowledge across diverse domains. For multi-hop reasoning, we utilized HotpotQA [31] and LongBench [32], which primarily test the model's ability to integrate information across multiple contexts and generate coherent reasoning outcomes.

**Sample Selection.** Following the settings in AoT [19], we constructed the evaluation sample sets as follows: the entire test set of GSM8K (1,319 samples) was used; for LongBench, we applied the merged MuSiQue [33] and WikiMultiHopQA [34] subsets (a total of 400 samples); for all other datasets, the first 1,000 samples from the respective test sets were selected for evaluation.

**Baselines.** We compare MGRS with a range of Chain-of-Thought-based methods, covering both classical prompting strategies and advanced reasoning frameworks.

The traditional prompting baselines include Chain-of-Thought (CoT) [2], which guides the model to reason step by step; CoT with Self-Consistency (CoT-SC) [5], which improves answer reliability through multiple sampled reasoning paths and majority voting; Self-Refine [26], which iteratively improves its own outputs through self-correction. We also compare with several state-of-the-art reasoning frameworks. Among them, AFlow [35] adopts an agent-based workflow to dynamically coordinate the reasoning process. Forest of Thought (FoT) [16] integrates multiple reasoning trajectories for efficient problem solving. Atom of Thoughts (AoT) [19] decomposes reasoning into atomic units and organizes them into a directed acyclic structure, enabling flexible state transitions and faster convergence. All experiments employ the GPT-4o-mini model to ensure fairness and comparability across methods. For MGRS, we set the number of reasoning branches to 2 and the number of samples per branch to 4.

Table 2: Comparison of Reasoning Model Performance on MATH, BBH and HotpotQA tasks (%).

| Model | MATH | BBH | HotpotQA |
|---|---|---|---|
| Doubao-1.5-lite | 77.6 | 76.2 | 69.7 |
| Gpt5-nano | 91.9 | 90.4 | 72.9 |
| o3-mini | 92.4 | 91.1 | 76.2 |
| Avg. | 87.3 | 85.9 | 72.9 |
| CoT | | | |
| Doubao-1.5-lite | 79.3 | 77.8 | 72.1 |
| Gpt5-nano | 94.5 | 92.6 | 75.7 |
| o3-mini | 93.7 | 92.9 | 77.5 |
| Avg. | 89.2 | 87.8 | 75.1 |
| MGRS | | | |
| Doubao-1.5-lite | 84.1 | 83.5 | 78.7 |
| Gpt5-nano | 97.6 | 95.3 | 81.3 |
| o3-mini | 98.2 | 97.5 | 84.6 |
| Avg. | 93.3 | 92.1 | 81.5 |

Table 3. Performance Comparison on Game of 24 (%). Our method achieved the highest average ranking across different inference frameworks with markedly reduced LLM call overhead.

| Method | LLM invoked | Success |
|---|---|---|
| CoT | 1.0 | 4.4 |
| CoT-SC | 10.0 | 4.4 |
| GoT(k=1) | 7.0 | 5.3 |
| AoT | 4.0 | 11.6 |
| ToT | 13.7 | 74.0 |
| BoT | 3.0 | 82.4 |
| XoT | 1.8 | 85.4 |
| FoT(n=4) | 20.6 | 93.7 |
| MGRS (Ours) | 7.3 | 100.0 |

Table 4. Runtime Comparison of FoT and MGRS (h).

| Method | FoT | MGRS |
|---|---|---|
| Runtime | 12.2 | 0.9 |

**Results.** Table 1 shows the experimental results on the six benchmarks. As shown in the Table, MGRS consistently demonstrates stable and significant performance gains across various reasoning tasks. In terms of mathematical reasoning, MGRS achieves the accuracy of 86.8% on MATH dataset and 96.5% on GSM8K dataset, outperforming the state-of-the-art AoT by 3.2% and 1.5%, respectively. These results also surpass FoT (82.5%/94.0% on MATH/GSM8K) and other baselines (best: 83.0%/93.5% from AFlow). For multiple-choice reasoning, MGRS reaches 86.2% accuracy on BBH dataset, performing comparably to AoT (86.0%), while outperforming FoT (82.4%), AFlow (76.0%) and other methods. On MMLU-CF dataset, MGRS achieves a score of 74.2%, exceeding AoT and FoT by 3.2% and 4.1%, respectively. A similar trend is observed in multi-hop reasoning tasks. MGRS outperforms AoT by 2.2% on HotpotQA (82.8% vs. 80.6%) and by 1.5% on LongBench (69.9% vs. 68.8%), outperforming all other competitive baselines (best: 73.5%/61.0% from AFlow). Collectively, these results confirm that the design refinements in MGRS directly contribute to its superior performance.

Table 2 presents the performance of Direct Inference, CoT, and MGRS across three datasets: MATH, BBH, and HotpotQA. We employ three lightweight models: Doubao-1.5-lite, GPT5-nano, and o3-mini—as base models for evaluation.

As shown in the table, MGRS consistently outperforms both Direct Inference and step-by-step reasoning (CoT) across all three base models. On the MATH dataset, MGRS achieves an average accuracy of 93.3%, surpassing CoT (89.2%) and Direct Inference (87.3%). On HotpotQA, MGRS attains 81.5% accuracy, corresponding to relative improvements of 6.4% and 8.6% over CoT and Direct Inference, respectively. These results demonstrate the effectiveness of MGRS in enhancing reasoning accuracy across diverse task types. Moreover, MGRS exhibits strong adaptability to different model architectures. When integrated into o3-mini, the framework achieves the accuracies of 93.2%, 73.1%, and 81.6% on MATH, BBH, and HotpotQA, respectively—establishing a group of new performance benchmarks across all three datasets.

## 4.2. The game of 24

The Game of 24 requires using four given numbers exactly once to construct an arithmetic expression that evaluates to 24. This task has been widely adopted to assess the performance of tree-structured reasoning models. Following the experimental setup of Forest of Thought (FoT) [16], we use the same dataset containing 95 valid test cases and a consistent set of baseline methods, including Chain-of-Thought (CoT) [2], CoT with Self-Consistency (CoT-SC) [5], Tree-of-Thoughts (ToT) [3], Graph-of-Thoughts (GoT) [4], Buffer-of-Thoughts (BoT) [36], and Everything-of-Thoughts (XoT) [15]. In addition, we incorporate Forest of Thought (FoT) [16] and further implement a customized version of Atom of Thoughts (AoT) [19] specifically adapted for the Game of 24 evaluation.

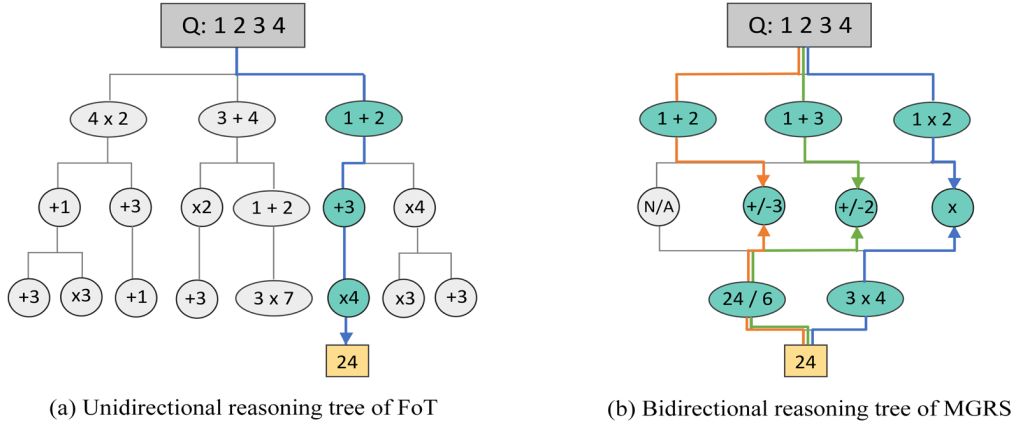

Figure. 2. Comparison between the reasoning tree of FoT and the bidirectional reasoning tree of MGRS. In the Game of 24 experiment, MGRS employs two type of reasoning branches to solve the problem from both forward and backward directions.

For model configuration, all experiments are conducted using Mistral-7B-Instruct. The FoT baseline adopts a breadth-first search (BFS) implementation with a branching factor of four ($n = 4$). In MGRS, two reasoning branches are initialized to solve the problem alternately from opposite directions (forward and backward), as illustrated in Figure 2. If both branches converge at the same intermediate node, the final answer is immediately obtained; otherwise, we discard the backward reasoning path and continue with the standard forward reasoning process. The overall performance comparison is presented in Table 3.

As shown in Table 3, the performance comparison on the 24-point game clearly demonstrates the effectiveness of the proposed MGRS method in mathematical and logical reasoning tasks. The experimental results show that MGRS achieves the highest success rate of 100%, significantly outperforming FoT ($n = 4$) with 93.7%, as well as other tree-structured reasoning methods (best: 85.4% from XoT). In particular, AoT performs slightly worse on this task due to its lack of multi-branch exploration capability within a tree structure.

Compared with FoT, our method introduces an additional backward reasoning process, which enhances the diversity of reasoning paths and thereby improves the overall success rate. Notably, since the two reasoning branches are designed to intersect, the additional backward path not only avoids increasing the total reasoning time but actually results in a substantial reduction. This observation can be intuitively explained: MGRS constructs a forward reasoning tree of depth three and a backward reasoning tree of depth one, while the matching between forward and backward reasoning results is automatically handled by FoT's result verification function, requiring no extra LLM calls. When the forward and backward trees converge at the same intermediate node, the solution is immediately obtained and the subsequent forward reasoning process is skipped, leading to a significant reduction in overall computation time. As shown in Table 4, MGRS completes all test cases in 0.9 hours, achieving a 13.6× speedup over FoT. These results strongly demonstrate that introducing diverse and particularly intersecting reasoning branches can substantially improve computational efficiency.

### 4.3. Ablation study

The previous experiments have demonstrated the superior performance of MGRS across multiple tasks. To further examine the contribution of each component to the overall performance, we conduct an ablation study on the GSM8K dataset using GPT-4o-mini as the backbone model. We first investigate the effects of differentiated reasoning branches and repeated sampling on performance improvement. Specifically, we vary the number of reasoning branches ($n_b = 1, 2, 4$) and the number of samples per branch ($n_s = 1, 2, 4, 8$), and evaluate all combinations of these parameters. The results are illustrated in Figure 3.

As shown in Figure 3, increasing both $n_b$ and $n_s$ consistently improves performance. When $n_b = 4$ and $n_s = 8$, MGRS achieves its peak accuracy of 97.3%. Increasing the sampling size while fixing the number of branches leads to steady gains, though improvements gradually saturate. This indicates that additional sampling enhances generation stability but cannot surpass the model's intrinsic reasoning limit, resulting in diminishing returns at higher sampling levels.

Differentiated branching also contributes notably to performance improvement. The configuration with $n_b = 2$ clearly outperforms that with $n_b = 1$, while further increasing $n_b$ yields smaller incremental gains. We attribute this to two interacting factors: (1) integrating reasoning paths with diverse perspectives enables complementary inferences that enhance reasoning accuracy; and (2) in the GSM8K mathematical reasoning task, the solution space is relatively narrow, so excessive differentiation may cause the model to generate spurious or contradictory reasoning steps, negatively affecting the final results. We hypothesize that for more open-ended tasks with diverse reasoning trajectories, larger $n_b$ values may be more beneficial.

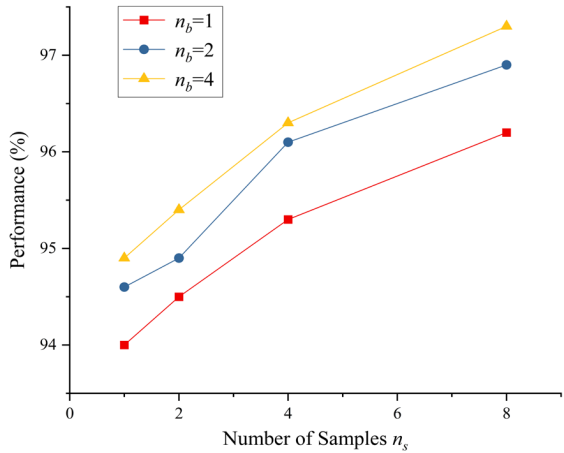

Figure. 3. Benefit analysis of MGRS: the return on the growth of the number of branches and samples.

Table 5: Ablation Study on MGRS Components (%).

| Method | BBH | MMLU-CF |
|---|---|---|
| MGRS (Original) | 85.8 | 74.7 |
| - Success Rate Estimation | 84.6 | 72.9 |
| - Cross & Self Validation | 84.8 | 73.0 |
| - DAG Construction | 84.3 | 73.6 |

In the second set of experiments (Table 5), we evaluate the effectiveness of the three core components of MGRS. Removing the success rate estimation module leads to performance drops of 1.2% and 1.8% on the two evaluation tasks, respectively. This degradation occurs because, without success rate estimation, reasoning paths across hierarchical layers are treated as equally reliable. Lower-level paths are typically more numerous but less coherent, increasing the likelihood of incorrect answers. Similarly, removing the combined cross- and self-validation mechanism results in an average performance decline of 1.4%, demonstrating that the composite validation effectively integrates and refines multi-branch reasoning paths. Finally, eliminating the Directed Acyclic Graph (DAG) construction process also degrades performance, which we attribute to the implicit atomization of reasoning chains introduced during DAG construction, providing additional granularity and stability in reasoning.

## 5. Conclusion

This work presents the Multi-chain Graph Refinement and Selection (MGRS) framework. MGRS inherits the strengths of existing chain-of-thought architectures such as step-by-step reasoning and DAG-based topologies while introducing several key innovations, including differentiated branching, integrated cross- and self-verification, and success-rate-based answer selection. Through these specific designs, MGRS establishes a new paradigm for graph-based reasoning frameworks. Extensive experiments across multiple benchmarks demonstrate that MGRS consistently enhances performance across diverse task types and difficulty levels, validating its effectiveness in improving the reasoning capability of large language models while maintaining computational efficiency.

## 6. Limitation

A key limitation of MGRS lies in its reliance on carefully crafted manual prompts for constructing differentiated reasoning branches. Forcing the model to generate explicitly diverse reasoning paths may sometimes induce hallucinations, leading to reduced accuracy. This issue is particularly pronounced in tasks with relatively uniform or single-solution reasoning patterns. We observe that incorporating a small amount of prior knowledge to guide the model in building task-specific differentiated branches can effectively mitigate this problem, although it requires prompt tuning tailored to the task type. This limitation highlights the need for future research to develop more effective guidance mechanisms that enable differentiated branches to provide accurate and diverse auxiliary reasoning signals while reducing the manual overhead associated with prompt design.

## References

[1] Z. Lin, T. Liang, J. Xu *et al.* Critical Tokens Matter: Token-Level Contrastive Estimation Enhances LLM's Reasoning Capability. *arXiv preprint arXiv:2411.19943,* 2024.
[2] J. Wei, X. Wang, D. Schuurmans *et al.* Chain-of-thought prompting elicits reasoning in large language models. *Advances in neural information processing systems,* vol. 35, pp. 24824-24837, 2022.
[3] S. Yao, D. Yu, J. Zhao *et al.* Tree of thoughts: Deliberate problem solving with large language models. *Advances in neural information processing systems,* vol. 36, pp. 11809-11822, 2023.
[4] M. Besta, N. Blach, A. Kubicek *et al.*, "Graph of thoughts: Solving elaborate problems with large language models," in *Proceedings of the AAAI conference on artificial intelligence*, 2024, vol. 38, no. 16, pp. 17682-17690.
[5] X. Wang, J. Wei, D. Schuurmans *et al.* Self-consistency improves chain of thought reasoning in language models. *arXiv preprint arXiv:2203.11171,* 2022.
[6] S. Huang, Z. Ma, J. Du *et al.* Mirror-consistency: Harnessing inconsistency in majority voting. *arXiv preprint arXiv:2410.10857,* 2024.
[7] N. Shinn, F. Cassano, A. Gopinath *et al.* Reflexion: Language agents with verbal reinforcement learning.

[8] S. Welleck, X. Lu, P. West *et al.* Generating sequences by learning to self-correct. *arXiv preprint arXiv:2211.00053,* 2022.
[9] H. Luo, N. Morgan, T. Li *et al.* Beyond Context Limits: Subconscious Threads for Long-Horizon Reasoning. *arXiv preprint arXiv:2507.16784,* 2025.
[10] T. Kojima, S. S. Gu, M. Reid *et al.* Large language models are zero-shot reasoners. *Advances in neural information processing systems,* vol. 35, pp. 22199-22213, 2022.
[11] L. Wang, W. Xu, Y. Lan *et al.* Plan-and-solve prompting: Improving zero-shot chain-of-thought reasoning by large language models. *arXiv preprint arXiv:2305.04091,* 2023.
[12] J. Long. Large language model guided tree-of-thought. *arXiv preprint arXiv:2305.08291,* 2023.
[13] D. Zhang, X. Huang, D. Zhou *et al.* Accessing gpt-4 level mathematical olympiad solutions via monte carlo tree self-refine with llama-3 8b. *arXiv preprint arXiv:2406.07394,* 2024.
[14] L. Kocsis and C. Szepesvári, "Bandit based monte-carlo planning," in *European conference on machine learning*, 2006: Springer, pp. 282-293.
[15] R. Ding, C. Zhang, L. Wang *et al.*, "Everything of thoughts: Defying the law of penrose triangle for thought generation," in *Findings of the Association for Computational Linguistics: ACL 2024*, 2024, pp. 1638-1662.
[16] Z. Bi, K. Han, C. Liu *et al.* Forest-of-thought: Scaling test-time compute for enhancing llm reasoning. *arXiv preprint arXiv:2412.09078,* 2024.
[17] Y. Yao, Z. Li, and H. Zhao. Beyond chain-of-thought, effective graph-of-thought reasoning in language models. *arXiv preprint arXiv:2305.16582,* 2023.
[18] Y. Yao, Z. Li, and H. Zhao, "GoT: Effective graph-of-thought reasoning in language models," in *Findings of the Association for Computational Linguistics: NAACL 2024*, 2024, pp. 2901-2921.
[19] F. Teng, Z. Yu, Q. Shi *et al.* Atom of thoughts for markov llm test-time scaling. *arXiv preprint arXiv:2502.12018,* 2025.
[20] X. Liang, S. Song, Z. Zheng *et al.* Internal consistency and self-feedback in large language models: A survey. *arXiv preprint arXiv:2407.14507,* 2024.
[21] W. Zhou, Q. Wang, M. Xu *et al.*, "Revisiting the self-consistency challenges in multi-choice question formats for large language model evaluation," in *Proceedings of the 2024 Joint International Conference on Computational Linguistics, Language Resources and Evaluation (LREC-COLING 2024)*, 2024, pp. 14103-14110.
[22] Z. Chu, J. Chen, Q. Chen *et al.* BeamAggR: Beam Aggregation Reasoning over Multi-source Knowledge for Multi-hop Question Answering. *arXiv preprint arXiv:2406.19820,* 2024.
[23] E. H. Jiang, H. Luo, S. Pang *et al.* Learning to Rank Chain-of-Thought: An Energy-Based Approach with Outcome Supervision. *arXiv preprint arXiv:2505.14999,* 2025.
[24] R. Zhao, X. Li, S. Joty *et al.* Verify-and-edit: A knowledge-enhanced chain-of-thought framework. *arXiv preprint arXiv:2305.03268,* 2023.
[25] Z. Ling, Y. Fang, X. Li *et al.* Deductive verification of chain-of-thought reasoning. *Advances in Neural Information Processing Systems,* vol. 36, pp. 36407-36433, 2023.
[26] A. Madaan, N. Tandon, P. Gupta *et al.* Self-refine: Iterative refinement with self-feedback. *Advances in Neural Information Processing Systems,* vol. 36, pp. 46534-46594, 2023.
[27] D. Hendrycks, C. Burns, S. Kadavath *et al.* Measuring mathematical problem solving with the math dataset. *arXiv preprint arXiv:2103.03874,* 2021.
[28] K. Cobbe, V. Kosaraju, M. Bavarian *et al.* Training verifiers to solve math word problems. *arXiv preprint arXiv:2110.14168,* 2021.
[29] M. Suzgun, N. Scales, N. Schärli *et al.*, "Challenging big-bench tasks and whether chain-of-thought can solve them," in *Findings of the Association for Computational Linguistics: ACL 2023*, 2023, pp. 13003-13051.
[30] Q. Zhao, Y. Huang, T. Lv *et al.*, "Mmlu-cf: A contamination-free multi-task language understanding benchmark," in *Proceedings of the 63rd Annual Meeting of the Association for Computational Linguistics (Volume 1: Long Papers)*, 2025, pp. 13371-13391.
[31] Z. Yang, P. Qi, S. Zhang *et al.*, "HotpotQA: A dataset for diverse, explainable multi-hop question answering," in *Proceedings of the 2018 conference on empirical methods in natural language processing*, 2018, pp. 2369-2380.
[32] Y. Bai, X. Lv, J. Zhang *et al.*, "Longbench: A bilingual, multitask benchmark for long context understanding," in *Proceedings of the 62nd Annual Meeting of the Association for Computational Linguistics (Volume 1: Long Papers)*, 2024, pp. 3119-3137.
[33] H. Trivedi, N. Balasubramanian, T. Khot *et al.* ♫MuSiQue: Multihop Questions via Single-hop Question Composition. *Transactions of the Association for Computational Linguistics,* vol. 10, pp. 539-554, 2022.
[34] X. Ho, A.-K. D. Nguyen, S. Sugawara *et al.* Constructing a multi-hop qa dataset for comprehensive evaluation of reasoning steps. *arXiv preprint arXiv:2011.01060,* 2020.
[35] J. Zhang, J. Xiang, Z. Yu *et al.* Aflow: Automating agentic workflow generation. *arXiv preprint arXiv:2410.10762,* 2024.
[36] L. Yang, Z. Yu, T. Zhang *et al.* Buffer of thoughts: Thought-augmented reasoning with large language models. *Advances in Neural Information Processing Systems,* vol. 37, pp. 113519-113544, 2024.